\documentclass[default,iicol,sn-jnl]{sn-jnl}% Style for submissions to Nature Portfolio journals
\usepackage{graphicx}%
\usepackage{multirow}%
\usepackage{amsmath,amssymb,amsfonts}%
\usepackage{amsthm}%

\usepackage{mathrsfs}%
\usepackage[title]{appendix}%
\usepackage{xcolor}%
\usepackage{textcomp}%
\usepackage{manyfoot}%
\usepackage{booktabs}%
\usepackage{algorithm}%
\usepackage{algorithmicx}%
\usepackage{algpseudocode}%
\usepackage{listings}%
\usepackage{cuted}
\usepackage{geometry}%
\usepackage{aasMacro}

\newcommand{\taun}{\tau}           % normalised time in [0,1]
\newcommand{\taustar}{\tau^\star}  % readout on the normalised axis
\newcommand{\tauc}{\tau_c}         % crossover marker on the normalised axis
\newcommand{\sigman}{\tilde{\sigma}}% normalised std (so that \sigman(1)=1)
\newcommand{\gn}{\tilde{g}}        % noise schedule: \gn^2 = d(\sigman^2)/d\tau

\newcommand{\lc}{\ell_c}
\newcommand{\rg}{r_g}
\newcommand{\Brms}{B_{\mathrm{rms}}}

\newcommand{\kappaS}{\kappa}        % spatial diffusion coefficient (position space)
\newcommand{\Dvel}{D_v}             % velocity-space diffusion constant

\newcommand{\Kernel}{\mathcal{T}}   % Transport kernel (distinct from kurtosis K)

\begin{document}
\title[Physics-Anchored Diffusion Models]{Generative Diffusion Surrogates with Analytical Variance Schedule}

\author*[1,2,3]{\fnm{Patrick} \sur{Reichherzer}}\email{p.reichherzer@princeton.edu}
\author[1]{\fnm{Gianluca} \sur{Gregori}}
\author[4,5]{\fnm{David~N.} \sur{Hosking}}
\author[1]{\fnm{Subir} \sur{Sarkar}}

\affil[1]{\orgdiv{Department of Physics}, \orgname{University of Oxford}, \orgaddress{\city{Oxford} \postcode{OX1 3PU}, \country{UK}}}
\affil[2]{\orgdiv{Department of Astrophysical Sciences}, \orgname{Princeton University},
          \orgaddress{\city{Princeton} \state{NJ} \postcode{08544}, \country{USA}}}
\affil[3]{\orgname{Max Planck Institute for Security and Privacy},
          \orgaddress{\city{Bochum} \postcode{44799}, \country{Germany}}}
\affil[4]{\orgdiv{Department of Applied Mathematics and Theoretical Physics}, \orgname{University of Cambridge}, \orgaddress{\city{Cambridge} \postcode{CB3 0WA}, \country{UK}}}
\affil[5]{\orgname{Gonville \& Caius College}, \orgaddress{\city{Cambridge} \postcode{CB2 1TA}, \country{UK}}}

\abstract{Stochastic transport describes physical systems in which an initially structured distribution spreads under unresolved forcing, scattering, or heterogeneous media. Useful surrogates for such systems should be probabilistic, time-resolved, and able to represent non-Gaussian distributional structure. Generative diffusion models, which corrupt data with Gaussian noise and learn a reverse flow back to structured states, have these properties. Their noise schedules, however, are usually chosen heuristically: image and audio generation---the canonical use cases---provide no physical clock. In transport, by contrast, the variance, or mean-square displacement, is often known from macroscopic theory or empirical scaling even when the full distribution is not. Here we prescribe the forward noising rate as the time derivative of this variance, turning generative time into a calibrated transport clock. The variance path is enforced by construction, while the learned score field represents how non-Gaussian structure inherited from entrance data is smoothed along that path, requiring no intermediate-time physical transport data. For ballistic-to-diffusive transport in turbulent plasmas, the surrogate matches test-particle distributions, reproduces the laboratory-measured variance scale, and tracks the simulated kurtosis evolution without schedule tuning, enabling calibrated emulation and likelihood-based inference.}

\maketitle

\section{Introduction}
Generative diffusion models---as widely used in AI image, audio and video generation---work by learning to undo noise. Their forward diffusion process corrupts training data by successive addition of random noise. A neural network is trained to reverse this corruption, so that a highly corrupted sample---even pure noise---can be transformed into a sample resembling the training data (reverse diffusion process). In this paper, we use the same machinery to model physical systems that evolve by the physical accumulation of random increments, i.e., stochastic transport, in which an initially organized distribution is acted on by unresolved fluctuations such as turbulent magnetic fields, scattering centers, porous heterogeneities, molecular collisions, or random forcing.

A central design choice in diffusion models is the forward noise schedule \( g(\tau)\), which specifies the amount of noise added in the forward diffusion process as a function of a time-like parameter $\tau$ \citep{Ho2020,Song2021SDE,NicholDhariwal2021,Karras2022EDM}. For the variance-exploding (VE) formulation used in this work, the forward corruption is
\begin{equation}
    d \mathbf{x}_{\taun} \;=\; \gn(\taun)\, d\mathbf{W}_{\taun},\label{eq:ve_forward_main}
\end{equation}
where \(\mathbf{W}_{\taun}\) is a standard Wiener process with independent increments. If \(\sigman^2(\taun)\) denotes the accumulated post-entrance variance increment of the surrogate, then
\begin{equation}
\sigman^{2}(\taun)
=
\int_0^{\taun}\gn^{2}(u)\,du,
\qquad
\gn^{2}(\taun)
=
\frac{d}{d\taun}\sigman^{2}(\taun).
\label{eq:schedule_anchor}
\end{equation}
Thus, a nondecreasing macroscopic variance law directly determines the instantaneous noising rate of the diffusion model.

In image, audio and video generation, $g(\tau)$ is usually chosen heuristically (e.g., linear or cosine in $\tau$ \citep{Ho2020, NicholDhariwal2021}), and tuned to maximize output quality or likelihood. For stochastic transport, by contrast, the variance, interpreted as the mean-square displacement of transported particles, is often known more directly than the full distribution, either from macroscopic transport theory or from measured scaling laws. 
Examples include non-Fickian transport in porous media, modeled via fractional kinetics \citep{Berkowitz2000WRR,Edery2015PRE}; intracellular subdiffusion in crowded cytoplasm \citep{Weiss2004BiophysJ,GoldingCox2006PRL}; superdiffusive dispersion in turbulence (Richardson's law) \citep{Richardson1926}; and charged-particle transport in stochastic magnetic fields. Analogous scaling laws also arise in broader stochastic-spreading problems such as human mobility and rough volatility \citep{Brockmann2006Nature, GatheralJaissonRosenbaum2014}.  Treating~$\tau$ as a normalized physical time, such a law can be written as \(\sigman^2(\taun)\) and used to calibrate the diffusion-model clock at the level of the variance.

\begin{figure}[t!]
\centering
\includegraphics[width=0.93\columnwidth]{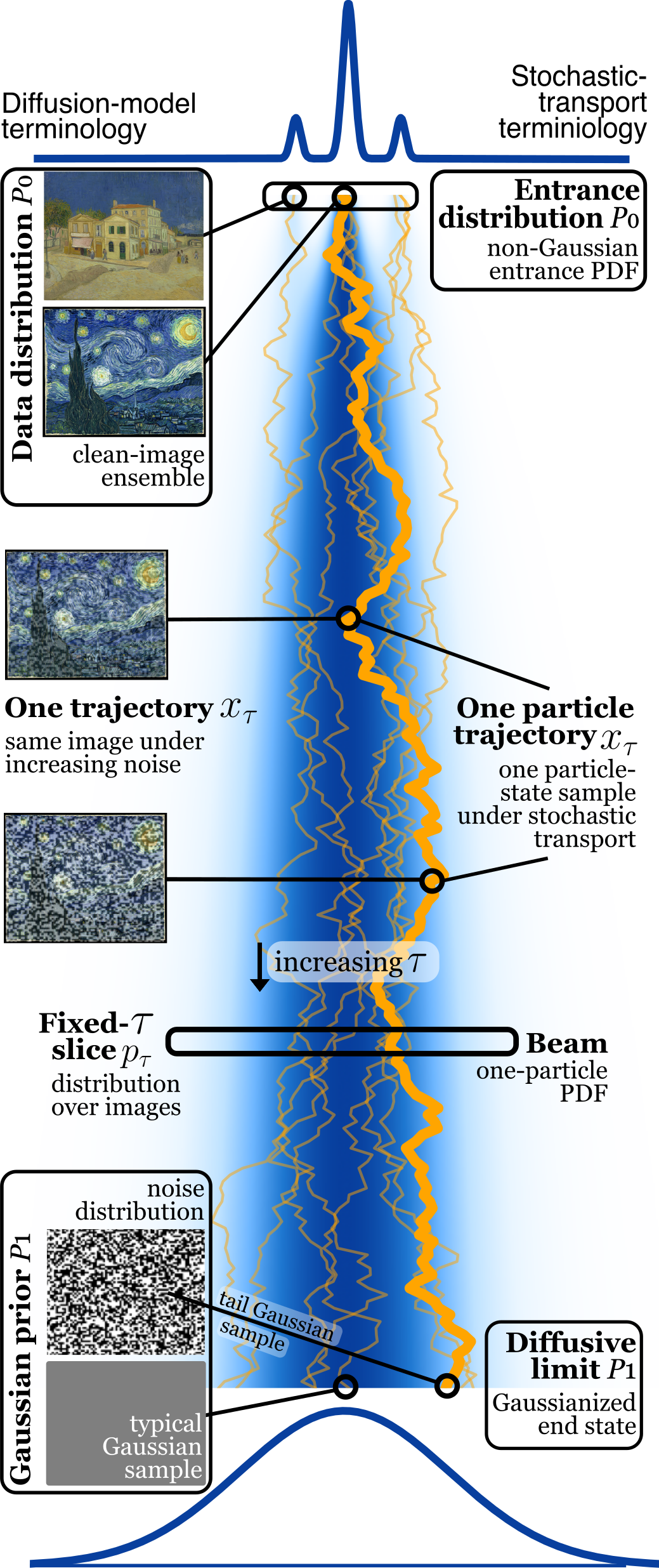}
\caption{\textbf{Diffusion–transport terminology map.} Individual curves show sample trajectories $x_\tau$; the blue density field shows the evolving marginal $p_\tau$. A fixed-$\tau$ slice is interpreted as a distribution over noised images in diffusion modeling and as the beam one-particle probability density function in stochastic transport, connecting the entrance/data distribution $P_0$ to the Gaussianized limit $P_1$.}
\label{fig:fig1}
\end{figure}

In this work, we therefore consider generative diffusion models with the noise schedule \(\gn(\taun)\) anchored to a known physical variance law \(\sigman^2(\taun)\) as surrogates for stochastic transport. Anchoring the variance separates the scale dynamics from the remaining distributional shape: the variance is prescribed analytically, while the learned reverse model represents higher-order structure, including non-Gaussian structure inherited from the entrance data. This calibration fixes the ensemble spreading at the variance level, but not the underlying microscopic dynamics: the synthetic Gaussian noising process does not, by itself, impose finite propagation speed, memory effects, or trajectory correlations. Figure~\ref{fig:fig1} summarizes the mapping between diffusion-model and stochastic-transport terminology used throughout this work.

This variance-clock construction differs from existing uses of generative models in scientific surrogate modeling. Diffusion models are increasingly used as scientific surrogates \citep{Thuerey2021,greif2023physics} including to synthesize particle trajectories from test-particle simulations \citep{2025ApJS..277...48M}, but their generative timescale is typically algorithmic, where the forward schedule is chosen or tuned for objectives such as sample quality, likelihood, or sampling efficiency~\citep{Ho2020,Song2021SDE,kingma2021variational,Dockhorn2021CritDamp,Bao2022AnalyticDPM,Karras2022EDM}. A separate line of scientific machine learning incorporates physical information by penalizing residuals of governing equations or imposing other physics-based constraints \citep{Raissi2019PINN,Thuerey2021,Karniadakis2021Review,Bastek2024PhysInformed}. Path-based generative methods, including flow matching, stochastic interpolants, and Schr\"odinger bridges, learn vector fields or stochastic processes along prescribed probability paths between distributions \citep{Lipman2023FM,Albergo2023StochInterpolants,DeBortoli2021,Duong2025Kac,Baldan2025PBFM}. These paths need not coincide with the physical transport clock, although trajectory data or physical interpolants can be incorporated in scientific applications \citep{WaldSteidl2025}. Other approaches solve inverse problems by approximately reversing a simulator and correcting the result with a learned score~\citep{Holzschuh2023}.

Operationally, we modify only the scalar schedule in the standard VE-SDE framework \citep{Song2021SDE}. The schedule $\gn^2(\taun)$ is derived directly from the macroscopic variance law $\sigman^2(\taun)$, calibrating the generative clock to physical time at the level of the second moment. This preserves existing diffusion-model infrastructure while enabling analytic moment tracking.
It also permits entrance-only training: the network is trained using entrance samples and their synthetic corruptions along the prescribed variance path without requiring physical transport data at intermediate times. Sampling at a prescribed readout time \(\taustar\) therefore gives the surrogate marginal associated with \(\sigman^2(\taustar)\).

In the following, we develop a framework for integrating macroscopic transport laws into generative diffusion models. We provide the theoretical basis in~\S\,\ref{sec:results_framework}, apply the method to telegraph-type ballistic-to-diffusive transport in~\S\,\ref{sec:results_finitespeed}, demonstrate entrance-only training in~\S\,\ref{sec:results_capabilities}, and use the learned model as a differentiable surrogate for laboratory plasma transport in~\S\,\ref{sec:results_lab}. In~\S\,\ref{sec:intermittency}, we apply the same construction to heavy-tailed and intermittent transport, where early-time transport statistics are strongly non-Gaussian. The resulting surrogate is indexed by physical time and is suitable for calibrated scientific emulation and likelihood-based inference.

\section{Results}\label{sec:results}

\subsection{Central Limit Theorem motivates physics-anchored
Gaussian surrogates}\label{sec:results_framework}

In many physical systems, macroscopic transport arises from the accumulation of microscopic perturbations, such as scattering events or stochastic collisions, that are correlated over a finite length scale~$\lc$ or, equivalently, a correlation time~$\tauc$.  Below this scale, the microphysics may be complex and strongly non-Gaussian. If coarse-grained perturbations separated by $\lc$ are weakly dependent and have finite second moment, the central limit theorem (CLT) motivates a Gaussian approximation to the accumulated transport kernel at scales large compared with~$\lc$.  After $N\!\sim\!\taun/\tauc$ approximately decorrelated contributions on the normalized clock with finite fourth moment, standardized higher cumulants relax; in particular, the excess kurtosis of the accumulated kernel falls as~$1/N$ (Methods, \S\,\ref{sec:error_quantification}).

We use this Gaussianization as a coarse-grained modeling assumption. Sub-$\lc$ non-Gaussian structure is represented through the entrance distribution~$P_0$, while subsequent post-entrance spreading is approximated by a Gaussian kernel whose variance is prescribed by the macroscopic transport law.  Thus non-Gaussian structure in the model marginals is inherited from~$P_0$ and then diluted by the prescribed variance growth, unless the physical transport kernel itself remains strongly non-Gaussian.  We exploit this construction in \S\,\ref{sec:intermittency} to propagate intermittency from sub-$\lc$ field statistics, while the anchored variance law supplies the scale evolution.

The Gaussianized spreading above~\(\lc\) is encoded by the accumulated post-entrance variance~\(\sigman^2(\taun)\). If a macroscopic law is written for the total variance, then \(\sigman^2(\taun)\) should be read as the variance increment relative to the entrance variance. We implement this variance increment using the VE process introduced in Eqs.~\eqref{eq:ve_forward_main}--\eqref{eq:schedule_anchor}, now initialized with
\[
    \mathbf{x}_0\sim P_0 .
\]
With \(\mathrm{Cov}[d\mathbf{W}_{\taun}]=I\,d\taun\), the resulting
marginal law at each fixed \(\taun\) can equivalently be written as
\[
    \mathbf{x}_{\taun}
    =
    \mathbf{x}_0+\sigman(\taun)\boldsymbol{\epsilon},
    \qquad
    \boldsymbol{\epsilon}\sim\mathcal N(0,I).
\]
Each synthetic Gaussian increment represents the coarse-grained, decorrelated part of the transport only at the level of the surrogate. The Markov approximation amounts to neglecting memory beyond~\(\tauc\) in these coarse-grained increments. The marginal distribution $p_{\taun}(\mathbf{x})$ generated by Eq.~\eqref{eq:ve_forward_main} is therefore the convolution of~$P_0$ with a Gaussian kernel (Methods, \S\,\ref{sec:gauss_surrogate}).  Starting from~$P_0$, each state accumulates independent Gaussian displacement with per-coordinate variance~$\sigman^2(\taun)$:
\begin{equation}
p_{\taun}(\mathbf{x})
  = (P_0 * \mathcal{N}(0,\sigman^2(\taun)I))(\mathbf{x}).
\label{eq:gaussian_convolution}
\end{equation}
Thus the total per-coordinate variance of the model marginal is $\sigma_0^2+\sigman^2(\taun)$, where $\sigma_0^2=\mathrm{Var}(P_0)$ denotes the entrance variance.

The VE process therefore acts as a Gaussian surrogate for the post-entrance transport kernel.  When the physical transport kernel is centered and isotropic, this Gaussian is the moment-matched member of the centered isotropic Gaussian family and minimizes cross-entropy within that family (equivalently, forward KL when the physical kernel has a density and finite differential entropy) (Methods,~\S\,\ref{sec:optimality}). The associated score is \(\mathbf{s}(\mathbf{x},\taun)=\nabla_{\mathbf{x}}\log p_{\taun}(\mathbf{x})\), which we approximate by a neural network \(\mathbf{s}_\theta\). Because the variance is enforced analytically by the schedule, this network is not required to learn the second-moment evolution. Instead, it represents the modeled evolution of the remaining distributional shape, including non-Gaussian structure inherited from~\(P_0\).  The trained score field makes the surrogate useful for inversion and likelihood-based inference (see~\S\,\ref{sec:results_capabilities}).

The Gaussian convolution~\eqref{eq:gaussian_convolution} yields a closed-form relaxation of entrance kurtosis.  We quantify tail heaviness using the excess kurtosis $K:=\kappa_4/\sigma^4$, where $\kappa_4$ is the fourth cumulant; a Gaussian has $K=0$, while heavy-tailed distributions have $K>0$. Let $K_0$ be the entrance excess kurtosis.  Because the Gaussian kernel contributes no fourth cumulant, the fourth cumulant of the model marginal is inherited from~$P_0$, while the variance grows as $\sigma_0^2+\sigman^2(\taun)$.  The model excess kurtosis therefore relaxes deterministically as
\begin{equation}
K_M(\taun)\;=\;K_0\left(\frac{\sigma_0^2}
  {\sigma_0^2+\sigman^2(\taun)}\right)^{\!2},
\label{eq:kurtosis_evolution_ve_main}
\end{equation}
entirely governed by the anchored variance path~$\sigman^2(\taun)$.

For centered transport, the mean and variance are fixed by construction. The residual mismatch between the physical kernel and the Gaussian surrogate therefore lies in unresolved distributional shape, including higher cumulants when they exist.  We quantify the leading mismatch relevant here using the kurtosis gap, defined in Methods, \S\,\ref{sec:error_quantification}, Eq.~\eqref{eq:kurt_gap}.

\subsection{Variance laws as noise schedules}
\label{sec:results_finitespeed}

In the coarse-grained picture of \S\,\ref{sec:results_framework}, $\tauc$ separates correlated from approximately decorrelated perturbations. For telegraph-type transport, the same relaxation scale separates two transport regimes. Here and below, $\tauc$ is expressed on the same normalized clock as $\taun$. Unlike ordinary Fickian diffusion, where the accumulated variance grows linearly ($\sigman^2\!\propto\!\taun$), many real systems show a ballistic-to-diffusive transition.

The two regimes have distinct signatures. At short times ($\taun \ll \tauc$), correlations persist: the direction or velocity has not yet decorrelated, so displacements grow coherently. This is the ballistic regime, where variance grows quadratically ($\sigman^2 \propto \taun^2$). Once $\taun \gg \tauc$, many approximately independent perturbations have accumulated and the variance grows linearly ($\sigman^2 \propto \taun$). This is the ordinary diffusive regime.

A minimal model for this crossover is the Goldstein--Kac, or telegraph, process \citep{Goldstein1951}. In one dimension, a particle moves at finite speed \(c\) while its direction decorrelates on a correlation time \(t_c\). In dimensional variables, the corresponding density \(q(x,t)\) satisfies, up to convention-dependent normalizations of \(t_c\),
\begin{equation}
    \frac{\partial^2 q}{\partial t^2}
    + \frac{1}{t_c}\frac{\partial q}{\partial t}
    =
    c^2\,\frac{\partial^2 q}{\partial x^2}.
\label{eq:telegraph_pde}
\end{equation}
The second-order time derivative yields finite-speed, persistent transport with ordinary diffusion as its long-time, large-scale limit.

In this work we do not use the full telegraph kernel as the VE surrogate kernel. We use only its mean-square displacement to define the variance clock. For a point source with symmetric initial velocities, the post-entrance variance increment has the form
\[
    \langle x^2(t)\rangle
    =
    2D\left[t-t_c\left(1-e^{-t/t_c}\right)\right].
\]
After normalizing physical time by \(t_{\rm obs}\), so that \(\taun=t/t_{\rm obs}\) and \(\tauc=t_c/t_{\rm obs}\), and rescaling the final accumulated variance to unity, this becomes
\begin{equation}
\sigman^{2}(\taun) \;=\; C^{-1} \left[ \taun -
  \tauc\!\left(1 - e^{-\taun/\tauc}\right) \right],
\label{eq:variance_telegraph_norm}
\end{equation}
where \(C = 1 - \tauc(1 - e^{-1/\tauc})\) enforces
\(\sigman^2(1)=1\).

Applying the anchoring
principle~\eqref{eq:schedule_anchor}, the VE noise schedule is
\begin{equation}
\gn^{2}(\taun) \;=\; \frac{d}{d\taun}\sigman^{2}(\taun)
  \;=\; C^{-1} \left[ 1 - e^{-\taun/\tauc} \right].
\label{eq:sde_forward_norm}
\end{equation}

\begin{figure}[t!]
\centering
\includegraphics[width=0.96\columnwidth]{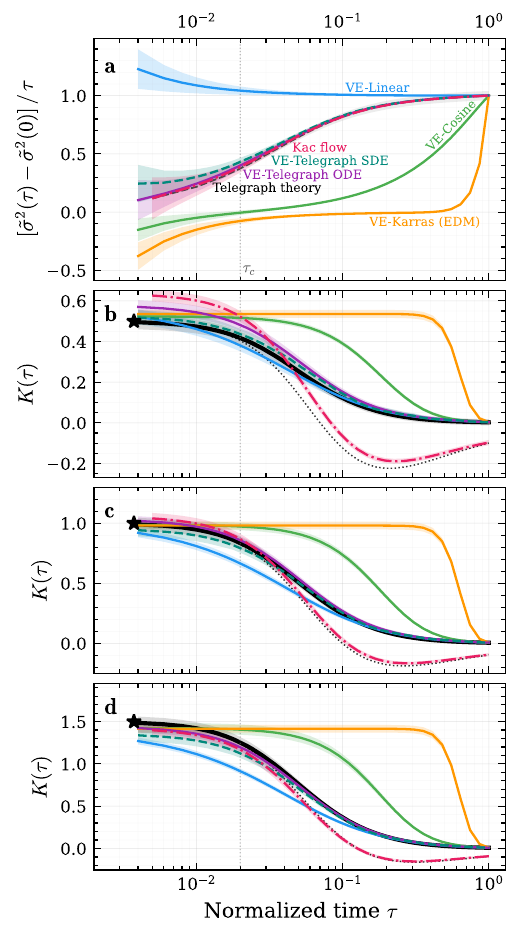}
\caption{\textbf{Schedule ablation under controlled training
budget (physics-anchored vs.\ heuristic schedules).}
\textbf{a)} Running diffusion coefficient $[\tilde{\sigma}^2(\tau)-\tilde{\sigma}^2(0)]/\tau$ measured from model samples as a function of normalized time $\tau$. Six different model configurations are shown: three heuristic schedules (Linear, Cosine, Karras/EDM), the Telegraph schedule (PF-ODE and reverse-SDE), and the Kac flow (a non-VE baseline). The analytical prediction of the Telegraph equation
(Eq.~\eqref{eq:variance_telegraph_norm}) is marked as a black dashed line.
\\\textbf{b--d)} Excess kurtosis $K(\tau)$ as functions of $\tau$ for three different entrance Student-$t$ distributions with $K_0\in\{0.5,1.0,1.5\}$ (black stars on the left edge). The analytical VE prediction is shown as a black line (Eq.~\eqref{eq:kurtosis_evolution_ve_main}); the dotted black line shows the Kac/telegraph reference computed from the analytical kernel moments (Eq.~\eqref{eq:kac_marginal}, Methods, \S\,\ref{sec:kac_kernel}). Implementation details are given in Algorithm~\ref{alg:pa_ve}. The standard deviations are shown as colored bands around the mean values.}
\label{fig:variance_kurtosis}
\end{figure}

A VE model equipped with this schedule has the prescribed ballistic-to-diffusive variance crossover by construction. Figure~\ref{fig:variance_kurtosis} tests whether this prescribed marginal path is realized after score training and reverse sampling, and how it differs from common heuristic schedules and from a finite-speed telegraph kernel.

Panel~\ref{fig:variance_kurtosis}a compares the accumulated variance of sampled models with the analytical telegraph prediction, plotted as a running diffusion coefficient. The black dashed curve is the telegraph variance law of Eq.~\eqref{eq:variance_telegraph_norm}; the colored curves are empirical variances measured from generated samples. VE-Telegraph ODE and VE-Telegraph SDE use the same anchored schedule, Eq.~\eqref{eq:sde_forward_norm}, but different reverse samplers: the deterministic probability-flow ODE and the stochastic reverse SDE. VE-Linear, VE-Cosine and VE-Karras/EDM use standard heuristic VE schedules under the same training budget. These schedules are valid generative choices, but their parameter \(\taun\) is not the telegraph physical clock, so they do not align with the timing and shape of the ballistic-to-diffusive crossover. The telegraph-anchored schedule is therefore the variance law used below for the ballistic-to-diffusive plasma application (Methods, \S\,\ref{sec:CR_transport} and \S\,\ref{sec:methods_mhdgrid}).

Panels~\ref{fig:variance_kurtosis}b--d test the higher-order statistics of the same sampled marginals. The solid black curve is the closed-form prediction of the Gaussian VE surrogate, Eq.~\eqref{eq:kurtosis_evolution_ve_main}; the colored curves are the excess kurtosis measured from trained sampled models for three entrance Student-\(t\) distributions. Agreement of the VE-Telegraph ODE and VE-Telegraph SDE curves with the solid line shows that the learned score and reverse samplers reproduce the intended Gaussian-convolution marginal path. This is the non-trivial part of the ablation: the variance law is imposed by the scalar schedule, but the kurtosis relaxation is not used as a training target. 

The dotted black curve gives a different reference: the variance-matched Kac/telegraph kernel derived in Methods, \S\,\ref{sec:kac_kernel}. This kernel corresponds to a simple particle-level model for the telegraph equation: a particle moves at finite speed and changes direction at Poisson-distributed times. Its parameters are chosen so that it has the same variance path as Eq.~\eqref{eq:variance_telegraph_norm}. The comparison therefore shows that distinct transport mechanisms can share the same variance law while differing in their propagators, approach to the diffusive limit, and memory \citep{LiangOh2026}. The Gaussian VE surrogate and the Kac kernel have the same mean-square displacement, but different higher moments. The Kac/telegraph kernel has bounded support \citep{Goldstein1951} and can transiently exhibit negative excess kurtosis. This is a controlled example of higher-cumulant mismatch between the physical kernel and the Gaussian VE surrogate. The departure is visible in Fig.~\ref{fig:variance_kurtosis}\textbf{b--d}, most clearly in panel~c, as the gap between the dotted (Kac) and solid~(VE) theory curves.

Table~\ref{tab:ablation_chi2} quantifies these comparisons across readout times and entrance kurtoses using a seed-normalized squared-deviation statistic $Q$. The telegraph PF-ODE and reverse-SDE samplers give the smallest deviations from the anchored VE reference. For the Kac flow, the small variance deviation confirms the imposed variance match, while the larger kurtosis deviation reflects the different higher-cumulant structure of the finite-speed kernel.

\begin{table}[t!]
\centering
\caption{\textbf{Physical-time alignment diagnostic for the schedule ablation (Fig.~\ref{fig:variance_kurtosis}).} For each moment we report
$Q=N_\tau^{-1}\sum_i(\bar y_i-y_i^{\rm ref})^2/ \max[s_i^2,0.01\langle s^2\rangle]$, where $\bar y_i$ and $s_i^2$ are the across-seed mean and variance. The statistic is evaluated at the available logarithmically spaced readout points in $\tau\in[3\times10^{-3},1]$ and averaged over 10 training runs and $K_0\in\{0.5,1.0,1.5\}$. Smaller values indicate closer alignment.}
\label{tab:ablation_chi2}
\begin{tabular}{l c c}
\toprule
Model configuration & $Q$ (variance) & $Q$ (kurtosis) \\
\midrule
VE-Linear $+$ PF-ODE       & $2.0\times10^{2}$ & $16$ \\
VE-Cosine $+$ PF-ODE       & $1.7\times10^{4}$ & $98$ \\
VE-Karras/EDM $+$ PF-ODE   & $1.3\times10^{5}$ & $2.6\times10^{2}$ \\
VE-Telegraph $+$ PF-ODE    & $0.34$            & $1.5$  \\
VE-Telegraph $+$ Rev-SDE   & $1.6$             & $1.1$  \\
Kac flow $+$ ODE           & $2.1$             & $74$ \\
\bottomrule
\end{tabular}
\end{table}

The anchoring principle~\eqref{eq:schedule_anchor} applies to any known, nondecreasing variance path, whether available in closed form or as a differentiable numerical representation. Anomalous diffusion is a natural extension: for power-law scaling $\sigman^2(\taun)\propto \taun^\alpha$, the schedule is $\gn^2(\taun)\propto \alpha\taun^{\alpha-1}$ \citep{MetzlerKlafter2000,Thiel2014sBm}. This includes subdiffusive scaling ($\alpha<1$), as in some CTRW models, and superdiffusive scaling ($\alpha>1$); fractional Brownian motion provides Gaussian examples with $\alpha=2H$. For $\alpha<1$, the schedule is integrable but singular at $\taun=0$, so practical implementations start from a finite $\taun_{\min}$ or regularize the schedule near the origin. Fractional telegraph models also admit closed-form variance expressions \citep{OrsingherBeghin2009}. The practical constraint is a finite second moment: ideal L\'evy flights are incompatible, so one anchors to a tempered or truncated law instead. In these cases, the VE construction should be interpreted as a variance-calibrated marginal surrogate, not as a pathwise representation of the underlying non-Markovian dynamics.

\subsection{Entrance-only training and differentiable
inference}
\label{sec:results_capabilities}

\begin{figure*}[!t]
\centering
\includegraphics[width=1.85\columnwidth]{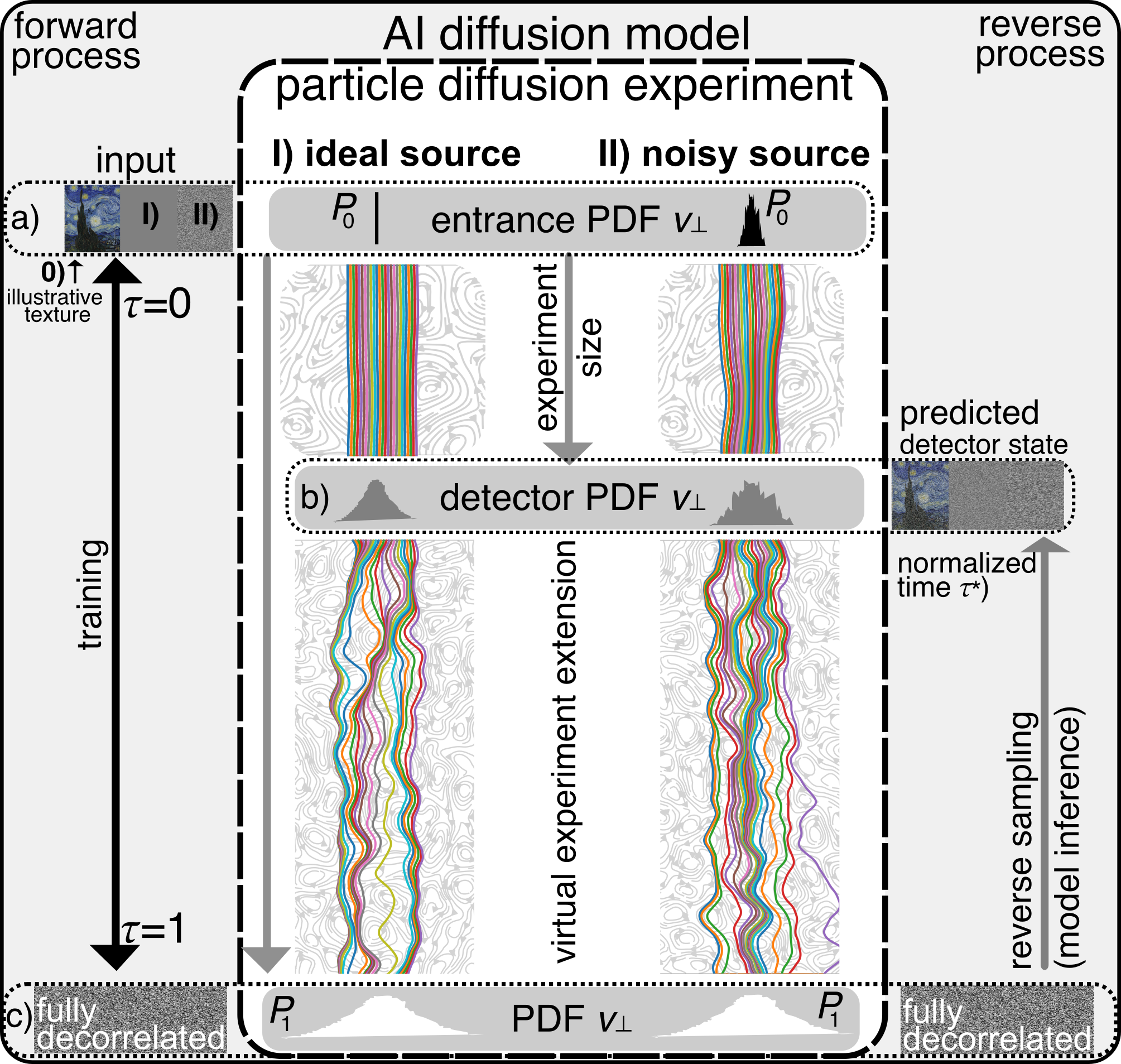}
\caption{\textbf{Physics-anchored diffusion maps experimental timescales to generative processes.}
\textbf{a}, Initial probability distribution $P_0$ at experimental source ($\taun=0$).
\textbf{b}, Target detector distribution $P_{\taustar}$ at physical readout $\taustar$ (e.g., one correlation time).
\textbf{c}, Fully decorrelated terminal state $P_1$ at normalized endpoint ($\taun=1$).
The forward physical process evolves $P_0\to P_{\taustar}$ in real time. The generative model trains on the full span $P_0\to P_1$ using the physics-anchored schedule $\gn^2(\taun)=d\sigman^2/d\taun$. The segment beyond the detector is a virtual extension of the experiment.
At inference, integrating the reverse probability-flow ODE from $P_1$ and halting at $\taustar$ recovers the detector-time marginal under the anchored surrogate clock without intermediate training data. In regimes where the transport kernel is near-Gaussian, this coincides with the physical detector distribution up to higher-cumulant mismatch.
This decoupling enables entrance-only training: the network learns shape evolution while the schedule analytically satisfies scale. Algorithm~\ref{alg:pa_ve} formalizes the training and readout-time sampling steps shown schematically in (a--c).
\label{fig:framework}}
\end{figure*}

Reverse sampling in a diffusion model must start from a terminal state in which the data structure has been washed out. In image generation, the forward corruption is simply run until this is the case. In our setting, the physical window closes earlier, at the detector readout $\taustar$, before the entrance structure is gone. We therefore extend the corruption beyond the detector along the same anchored variance law, until the accumulated variance dominates the entrance variance. This continuation, $\taun\in(\taustar,1]$, is the virtual extension of the experiment, as illustrated in Fig.~\ref{fig:framework}. It requires no data, because corrupting entrance samples to any $\taun$ needs only Gaussian noise of known variance. Because training and sampling share the same $\taun$-anchored marginals $p_\taun$, we integrate the PF-ODE backward from the fully decorrelated state $P_1$ (Fig.~\ref{fig:framework}c) and halt at $\taustar$, obtaining
$p_{\taustar}$ without intermediate data.

Given entrance samples, samples from $p_{\taustar}$ could also be produced by direct forward corruption.  The trained score field supplies what direct corruption alone does not: a differentiable transport map, PF-ODE likelihood estimates under the terminal-density assumptions specified in Methods, gradient-based parameter inference, and model-based backward propagation of observed detector ensembles to earlier times. The model is therefore best viewed as a stochastic propagator of distributions. 

The physics-anchored schedule ensures that the model marginals $p_{\taun}$ are indexed by physical time and have the prescribed accumulated transport variance $\sigman^2(\taun)$ at each~$\taun$. When the physical transport kernel is close to Gaussian, as expected after decorrelation over $\lc$, these anchored marginals provide a variance-exact approximation to the physical marginal. Remaining discrepancies lie in unresolved distributional shape, including higher cumulants when they exist, and are assessed here using the kurtosis gap (Methods, Eq.~\eqref{eq:kurt_gap}).

The training scheme does not require a known analytical expression for the forward convolution.  It requires only samples from a well-defined entrance distribution $P_0$ and the ability to add Gaussian corruptions with the prescribed variance.  For any positive corruption variance, Gaussian smoothing gives a regular model marginal, and denoising score matching targets the score of this full, generally non-Gaussian marginal distribution \citep{VincentEtAl2011,Song2021SDE}.

We verified entrance-only learning across entrance kurtoses using Student-$t$ families with fixed entrance variance.  As implied by Eq.~\eqref{eq:kurtosis_evolution_ve_main}, $K(\taun)$ preserves the entrance ordering and relaxes deterministically along the telegraph-anchored path (Fig.~\ref{fig:variance_kurtosis}b--d). When the physical kernel is approximately Gaussian, the model evolution matches the physical one at the level of the resolved marginal
statistics. Residual kernel non-Gaussianity is quantified by the kurtosis gap (Methods, Eq.~\eqref{eq:kurt_gap}).

The learned score field can also support parameter inference from detector data through the differentiable likelihood estimate obtained with the PF-ODE (Eq.~\eqref{eq:exact_likelihood_methods}). If $K_0$ is not known, it can be treated as a parameter of the entrance family $P_0(\,\cdot\,;K_0)$, equivalently by conditioning the score model on $K_0$ as described in Methods, \S\,\ref{sec:implementation}, and estimated by maximizing Eq.~\eqref{eq:exact_likelihood_methods} from an observation snapshot.

For the zero-drift VE process in Eq.~\eqref{eq:ve_forward_main}, the reverse-time SDE is
\[
d\mathbf{x}_{\taun}
=
-\gn^2(\taun)
\nabla_{\mathbf{x}}\log p_{\taun}(\mathbf{x}_{\taun})\,d\taun
+
\gn(\taun)d\bar{\mathbf{W}}_{\taun},
\qquad d\taun<0.
\]
The corresponding probability-flow ODE is
\[
d\mathbf{x}_{\taun}
=
-\frac{1}{2}\gn^2(\taun)
\nabla_{\mathbf{x}}\log p_{\taun}(\mathbf{x}_{\taun})\,d\taun,
\qquad d\taun<0,
\]
with the score replaced in practice by $\mathbf{s}_\theta(\mathbf{x}_{\taun},\taun)$.  In both forms, $\gn^2(\taun)$ ties the transport map to the prescribed variance law, while the network supplies the unresolved distributional structure. The learned score remains useful in settings such as inhomogeneous transport or nonlinear detector models, where analytic convolution is unavailable.

%%====================================================
\subsection{Application to turbulent plasma transport}\label{sec:results_lab}

\begin{figure*}[t]
\centering
\includegraphics[width=1.6\columnwidth]{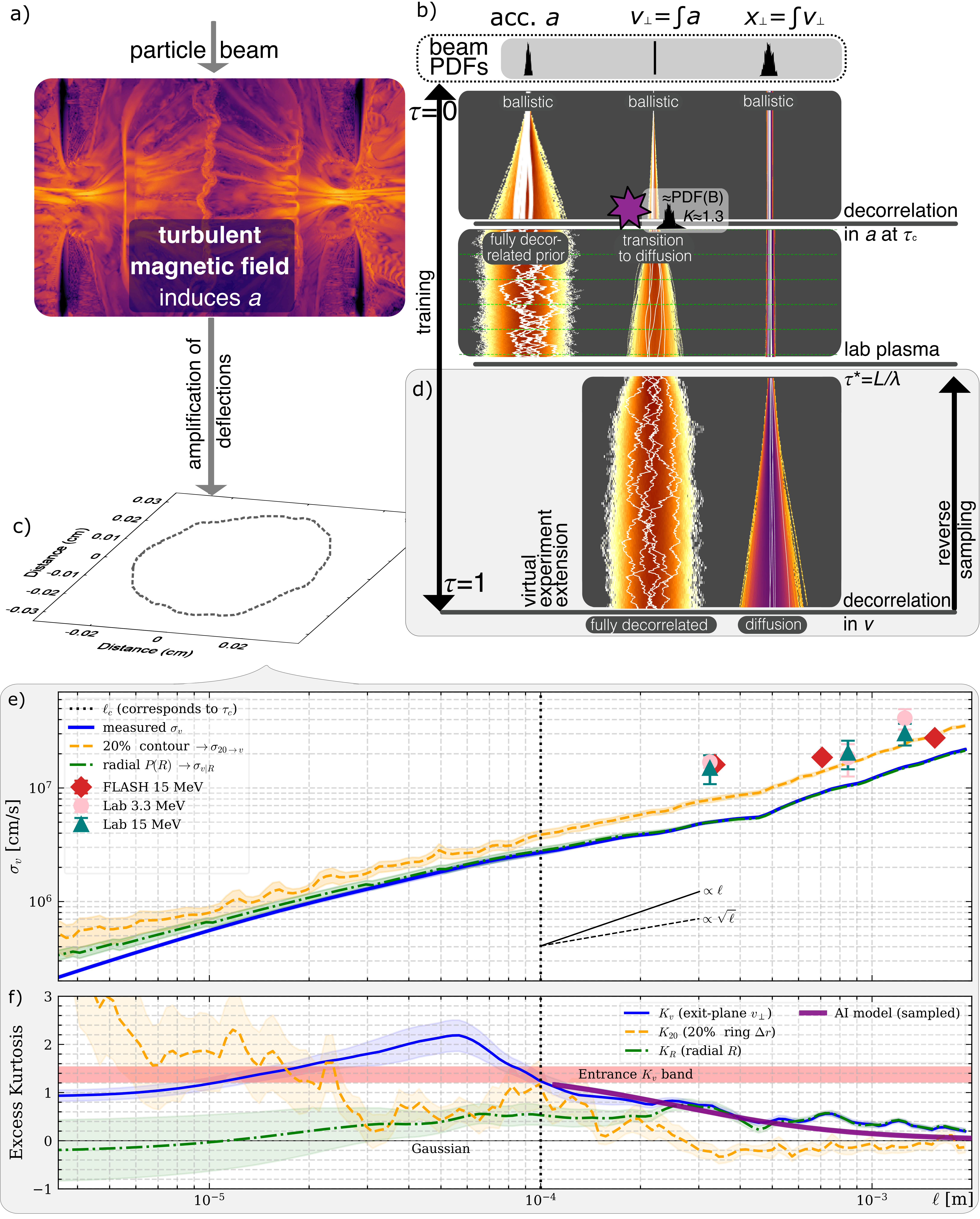}
\caption{\textbf{Physics‑anchored surrogate validation in turbulent plasmas.}
\textbf{a)} A narrow particle beam probes a magnetized turbulence snapshot, inducing stochastic transverse acceleration $a$.
\textbf{b)} Evolution of the PDFs of acceleration, velocity $v_\perp$, and displacement $x_\perp$.
\textbf{c)} Experimental diagnostic (proton radiography) (see \cite{Chen2020}).
\textbf{d)} Reverse PF‑ODE sampling visualization.
\textbf{e)} RMS transverse-speed magnitude $v_{\perp,\mathrm{rms}}$ versus physical path length $\ell$. The simulated variance increases super-diffusively at large $\ell$ because deflections then add up correlatedly due to the periodic continuation of the magnetic field.
\textbf{f)} Excess kurtosis of the signed transverse component $v_x$ versus $\ell$. Panel f contains test-particle and surrogate results only. Black points in panel e: experiment of \cite{Chen2020}. Colored bands/lines: test-particle simulation in the MHD field. The purple curve in panel f shows the sampled kurtosis evolution of the scalar anchored VE surrogate using the nearest trained single-component entrance-kurtosis condition. The surrogate captures the laboratory-measured variance scale and the simulated component-kurtosis evolution without schedule tuning. In panel e the model variance is not plotted separately because it is the imposed anchored variance path. Bands in \textbf{e} and \textbf{f} are bootstrap 68\% intervals resampled over the 20 independent simulation runs.} 
\label{fig:propagation_cartoon}
\end{figure*}

We now apply the physics-anchored VE model to the problem of charged-particle transport in turbulent magnetic fields, a scenario relevant for ultra-high-energy cosmic rays, solar energetic particles, and fusion plasmas. Since these systems are challenging to probe directly, scaled laboratory astrophysics experiments are used to investigate transport under controlled conditions \citep{2017JPlPh..83f9014B, 2018NatCo...9..591T, Chen2020}. The experimental technique, proton radiography (Fig.~\ref{fig:propagation_cartoon}a, c), uses a proton beam to probe the turbulence, measuring the resulting flux pattern $\Psi(\mathbf{r})$ (in contrast, we use $\overline{\Psi}(r)$ for the azimuthally averaged profile) at a detector. Here $\mathbf{r}=(x,y)$ are detector‑plane coordinates and $r=\|\mathbf{r}\|$ denotes radial distance.

Turbulent magnetic fields have a typical correlation length $\lc$ (the average size of a turbulent cell). We concentrate on the high-rigidity limit, in which the particle gyroradius $r_g=p/(|q|B)$ is much larger than the correlation length ($r_g\gg\ell_c$). In this limit, the physical picture is intuitive (Fig.~\ref{fig:propagation_cartoon}b): a high-energy particle traverses a turbulent cell (size $\lc$) quickly, giving rise to only a small angular deflection. The overall transport results from the accumulation of many such small-angle kicks across numerous independent cells. This scenario motivates the telegraph-type finite-speed transport framework (\S\,\ref{sec:results_finitespeed}). The physical time required to cross one cell defines \(t_c=\ell_c/v\), corresponding to the normalized correlation time \(\tauc=t_c/t_{\rm obs}\). The correlation length~$\lc$ plays the role of the coarse-graining scale of \S\,\ref{sec:results_framework} that determines the time~$\tauc$ needed to cross the coarse-graining scale. The ballistic-to-diffusive crossover of \S\,\ref{sec:results_finitespeed} therefore applies directly, with the telegraph variance law~\eqref{eq:variance_telegraph_norm} governing the transition. By anchoring the VE schedule to this law, the generative model is aligned with physical time at the level of the transport variance, without heuristic schedule tuning.

The nature of magnetic turbulence adds an additional complication (see~Methods, \S\,\ref{sec:methods_mhdgrid} for details). It is often intermittent, meaning the magnetic field exhibits rare, strong fluctuations rather than smooth variations. This results in heavy tails in the distribution of magnetic fields $P(B_\perp)$, characterized by positive excess kurtosis $K>0$. The particle transport also inherits these heavy tails due to rare large-angle scattering events. However, the choice of diagnostic determines whether these non-Gaussian features are observable (Fig.~\ref{fig:propagation_cartoon}e, f). A reliable experimental diagnostic is the $20\%$ contour of the flux, $C_{20}$, which measures the broadening of the beam core while suppressing the influence of the tails (see \cite[Appendix~C.2]{Chen2020} for details). This observable is Gaussian with $K\approx0$ because the beam radius is much larger than the transverse correlation length. The contour effectively averages over many independent turbulent patches. By the CLT (here applied to spatial averaging across the beam, distinct from the
temporal accumulation in~\S\,\ref{sec:results_framework}), this spatial averaging Gaussianizes the core statistics. For this specific observable, the moment-matched Gaussian surrogate is optimal within the class of Gaussian approximations (Methods,~\S\,\ref{sec:optimality}). In contrast, the azimuthally averaged profile $\overline{\Psi}(r)$ probes the full displacement PDF, including the tails, and remains sensitive to the intermittency inherited from the magnetic field statistics (Methods, \S\,\ref{sec:methods_kurtosis_relaxation}), as shown next.

\subsection{Modeling intermittency with the anchored VE
surrogate}
\label{sec:intermittency}

Section~\ref{sec:results_framework} introduced the modeling assumption used throughout the paper: sub-$\lc$ non-Gaussian structure is represented through the entrance distribution~$P_0$, while subsequent coarse-grained spreading is approximated by a Gaussian transport kernel with physics-anchored variance. We now give this construction a concrete physical realization.

In the high-rigidity regime of \S\,\ref{sec:results_lab}, a particle receives a small angular kick when crossing one turbulent cell. This kick is proportional to the path-averaged perpendicular magnetic field~$\overline{B}_\perp$ over a distance of order~$\lc$. For the scalar surrogate, we use one signed transverse component and represent $P_0$ by the trained Student-$t$ entrance condition whose kurtosis is closest to the value measured at the coarse-graining scale. Because the anchored surrogate is defined only for $\taun\ge\tauc$, the entrance statistic must be read at the corresponding path length $\ell=\lc$: we therefore take $K_0$ as the excess kurtosis of the signed transverse velocity component of the test-particle ensemble at $\ell=\lc$, giving $K_0 = 1.4\pm0.2$ (mean $\pm$ standard error over the 20 independent runs). This is the particle-level realization of the field statistic of Methods~\S\,\ref{sec:methods_kurtosis_relaxation}, where the linear kick relation gives $K(\Delta v_\perp(\tauc))=K(\overline{B}_\perp)$. This identification injects the non-Gaussian shape, including the heavy tails inherited from intermittent turbulence with $K_0\!>\!0$, at the coarse-graining scale~$\lc$, before subsequent multi-cell spreading. The physics-anchored telegraph schedule then fixes the variance growth and governs the relaxation toward the diffusive regime at the level of the second moment.

We validate this construction in Fig.~\ref{fig:propagation_cartoon}e and~f. The purple curve in Fig.~\ref{fig:propagation_cartoon}f shows the sampled kurtosis evolution predicted by the VE surrogate using the nearest trained entrance-kurtosis condition. The variance scale in Fig.~\ref{fig:propagation_cartoon}e is fixed by the telegraph-anchored schedule. Panel~f shows that the entrance statistics, propagated with the anchored variance law, capture the overall trend of the observed kurtosis evolution. In the diffusive regime, the empirical residual is $|K_{\mathrm{phys}}-K_M|=0.20\pm0.05$. The contribution from residual kernel non-Gaussianity is quantified by Eq.~\eqref{eq:kurt_gap} (Methods,~\S\,\ref{sec:methods_kurtosis_relaxation}).

\section{Discussion}\label{sec:discussion}
We have introduced a framework that links macroscopic physical transport laws directly to the variance schedule of a diffusion model. This physics-anchoring aligns the generative and physical timescales by construction, at the level of the variance. The variance path is enforced analytically. The score network is then needed only for the remaining non-Gaussian structure. Because training and sampling share the same $\tau$-anchored marginals, the reverse PF-ODE acts as a physics-timed denoiser: integrating backward from a detector ensemble at $\taustar$ recovers earlier surrogate marginals and—by stopping early—undoes only late-time noise. Along this reverse evolution the variance path and the kurtosis relaxation are preserved (Fig.~\ref{fig:variance_kurtosis}), so reconstruction is distributional (more detailed than moment inversion only).

The resulting Gaussian surrogate accurately describes diagnostics that average many independent contributions (e.g., the laboratory 20\% contour via the CLT). For Student-$t$ entrance distributions matched in variance and kurtosis, the VE model reproduces the predicted entrance-kurtosis relaxation (\S\,\ref{sec:intermittency}). The kurtosis gap quantifies the contribution from a non-Gaussian physical transport kernel.

This distinction is sharpest when comparing against the Kac/telegraph propagator. 
Since the test-particle marginals show positive excess kurtosis (Fig.~\ref{fig:propagation_cartoon}f), the data favor the Gaussian-kernel surrogate in this regime. The observed positive non-Gaussianity can be represented through the entrance statistics and does not require a bounded-support transport kernel.

\begin{figure}[t]
    \centering
    \includegraphics[width=1.04\columnwidth]{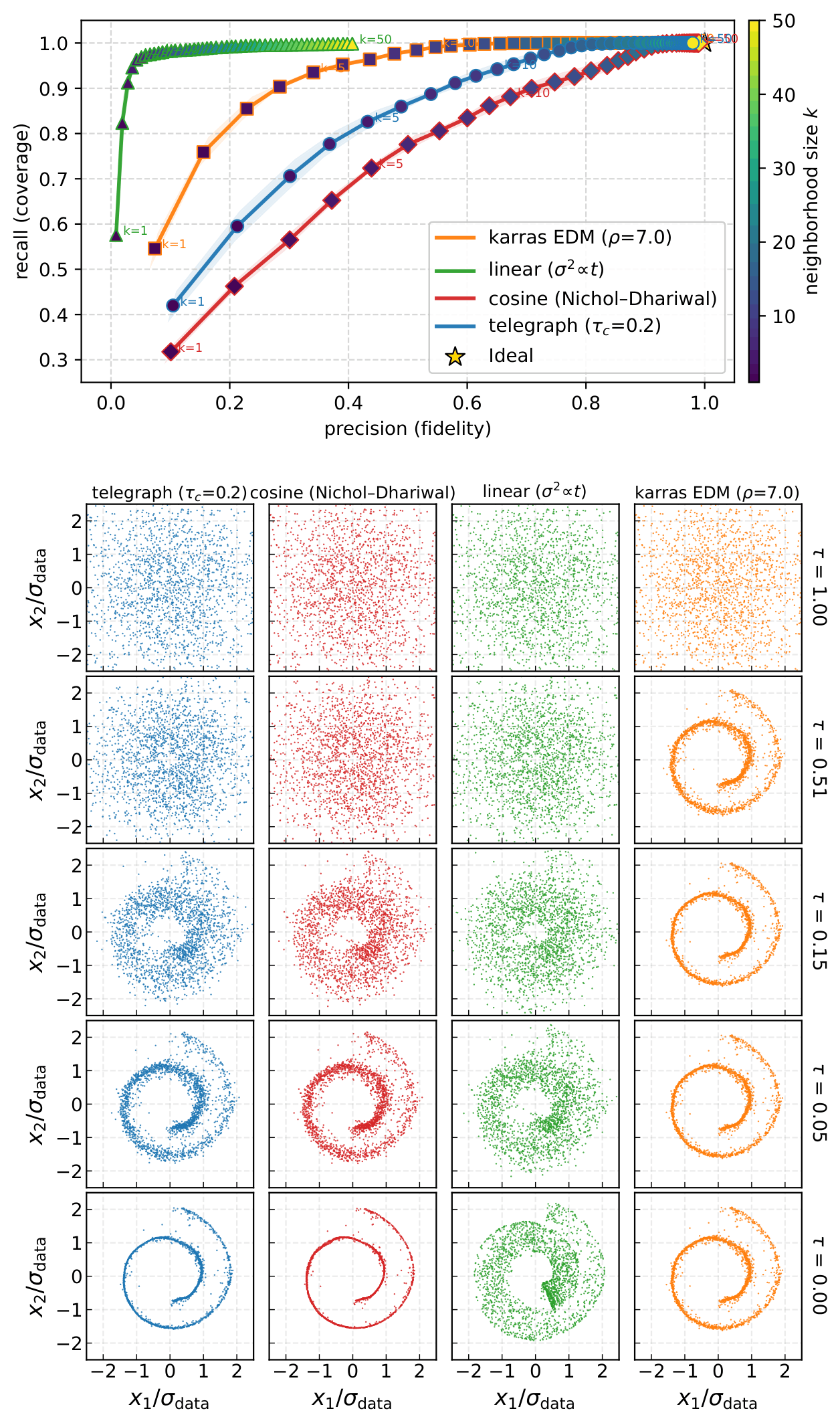}
    \caption{\textbf{Non-physics example.} \textbf{Top:} Precision–Recall curves on 2‑D swiss roll comparing telegraph ($\tauc=0.2$) to karras EDM, linear, and cosine schedules. Points along each curve are colored by neighborhood size $k$ in the nearest-center k-NN P/R diagnostic defined in Methods. Larger $k$ evaluates global structure. Shaded bands show $\pm 1\sigma$ over five independent sampling seeds. Telegraph provides a competitive precision-recall balance across the tested neighborhood sizes. \textbf{Bottom:} Examples of generated samples at different times. All curves share one model trained once with $\sigma$‑conditioning; we vary only the sampling path (solver, NFE, seeds, and noise bounds are identical), isolating the effect of the schedule.
}
    \label{fig:swiss_roll_pr}
\end{figure}

Because our contribution is a scalar path $\tilde\sigma^2(\tau)$, it can be used as a drop-in schedule in any VE sampler (e.g., Euler--Maruyama, predictor-corrector \citep{Song2021SDE}, Heun/second-order, or the EDM framework \citep{Karras2022EDM}). Figure~\ref{fig:variance_kurtosis} already demonstrates both PF-ODE and reverse-SDE sampling with the same schedule. In all physics results we train and sample on the same $\tau$‑anchored path to keep the generative clock calibrated to physical time at the level of the variance. In a non-physical Swiss-roll benchmark, the telegraph path provides a competitive precision--recall balance across the tested neighborhood sizes (Fig.~\ref{fig:swiss_roll_pr}; Methods, \S\,\ref{sec:methods_swiss_pr}). A theoretical explanation remains an open question.

Returning to the scientific setting, the learned score field provides differentiable likelihood estimates via the PF-ODE and can, in principle, support gradient-based inference for physical parameters entering the schedule or conditioning the score model. This capability extends to complex scenarios (e.g., heteroscedastic transport or nonlinear detectors) where analytic convolution does not exist, while naturally maintaining the correct variance scaling (Fig.~\ref{fig:propagation_cartoon}). Note that joint identifiability of multiple physical parameters requires either multiple temporal snapshots or additional constraints.

The variance-anchoring method we propose complements alternative techniques that learn explicit temporal variations in transport maps or trajectories, including flow matching \citep{Lipman2023FM, Baldan2025PBFM}, for which \cite{WaldSteidl2025} provides a unified overview, and reverse-time simulators \citep{Holzschuh2023}. Those approaches typically use trajectory data or interpolated pairs at multiple intermediate times to learn the complete transport map. Our focus on distributional snapshots allows for a highly data-efficient paradigm: training requires entrance samples and the prescribed variance law. This makes our method particularly suited to experimental conditions in which intermediate-time transport data are unavailable. The trade-off is that we predict the distribution of particles, not their individual paths.  The Kac flow baseline in Fig.~\ref{fig:variance_kurtosis} is itself a flow-matching model, providing a direct controlled comparison: it matches the telegraph variance but predicts sub-Gaussian intermediate marginals inconsistent with the observed positive kurtosis (Fig.~\ref{fig:propagation_cartoon}f). The combination of physically anchored schedules and path-learning networks is likely to lead to models which are compatible both with the macroscopic laws and the microscopic correlations. Learned or non-Brownian 1-D noising processes \citep{2025arXiv251012636C} offer another route to data-adapted schedules that could be combined with physics-anchored variance paths.

The application domains listed in the Introduction (anomalous diffusion, rough volatility, intracellular transport) are natural candidates. For processes with non-Gaussian kernels (e.g., CTRW propagators in porous media~\citep{Berkowitz2000WRR}), the leading fourth-order mismatch is quantified by the kurtosis gap~\eqref{eq:kurt_gap}. The entrance-statistics mechanism of \S\,\ref{sec:intermittency} captures leading non-Gaussian features. Wherever a system's variance growth is governed by a known law, that law can serve as the clock of a generative model. Such a distributional propagator, calibrated to physical time at the variance level, could serve as a transport module within a broader scientific world model. The network is then left to learn only what the law does not already know.

\newpage
\section{Methods}\label{sec:method}

Table~\ref{tab:notation} collects the principal symbols; each
is also defined at first use.

\begin{table}[t]
\centering
\caption{\textbf{Principal notation.}}
\label{tab:notation}
\begin{tabular}{@{}l l l l@{}}
\toprule
\textbf{Symbol} & \textbf{Meaning}
  & \textbf{Units} & \textbf{Defined} \\
\midrule
\multicolumn{4}{@{}l}{Generative model} \\[2pt]
$\taun$            & normalized time $\in[0,1]$
  & --- & \S\,1 \\
$\taustar$         & Readout (detector) time
  & --- & \S\,1 \\
$\sigman^2(\taun)$ & Acc. post-entrance variance, $\sigman^2(1)\!=\!1$
  & --- & Eq.\,\eqref{eq:schedule_anchor} \\
$\gn(\taun)$       & Noise schedule,
                     $\gn^2\!=\!d\sigman^2/d\taun$
  & --- & Eq.\,\eqref{eq:schedule_anchor} \\
$P_0$              & Entrance distribution
  & --- & Eq.\,\eqref{eq:ve_forward_main} \\
$p_\taun$          & Marginal at time $\taun$
  & --- & Eq.\,\eqref{eq:gaussian_convolution} \\
$\mathbf{s}_\theta$& Learned score
                     $\approx\nabla_{\mathbf{x}}\!\log p_\taun$
  & $[\mathbf{x}]^{-1}$ & \S\,\ref{sec:gauss_surrogate} \\
$K$                & Excess kurtosis $\kappa_4/\sigma^4$
  & --- & Eq.\,\eqref{eq:kurtosis_evolution_ve_main} \\
$\Kernel(\taun)$   & Physical transport kernel
  & --- & \S\,\ref{sec:optimality} \\
$\mathbf{W}_\taun$ & Standard Wiener process
  & --- & Eq.\,\eqref{eq:ve_forward_main} \\
\midrule
\multicolumn{4}{@{}l}{Physical transport} \\[2pt]
$\tauc$          & Correlation / crossover time
  & --- & \S\,\ref{sec:results_framework} \\
$\lc$            & Correlation length
  & $\mu$m & \S\,\ref{sec:results_framework} \\
$\rg$            & Gyroradius $p/(qB)$
  & $\mu$m & \S\,\ref{sec:results_lab} \\
$\Dvel$          & Velocity-space diffusion coeff.\
  & m$^2$\,s$^{-3}$ & \S\,\ref{sec:CR_transport} \\
$\kappaS$        & Spatial diffusion coefficient
  & m$^2$\,s$^{-1}$ & \S\,\ref{sec:CR_transport} \\
$\Brms$          & RMS magnetic field
  & kG & Table\,\ref{tab:baseline_params} \\
\bottomrule
\end{tabular}
\end{table}

\subsection{The VE framework and Gaussian convolution}
\label{sec:gauss_surrogate}

We adopt the standard VE framework \citep{Song2021SDE}.  The
forward SDE and resulting Gaussian convolution are given in
Eqs.~\eqref{eq:ve_forward_main} and~\eqref{eq:gaussian_convolution};
here we record the training objective.
We set $\gn^2(\taun)=\frac{d}{d\taun}\sigman^2(\taun)$ so
that, by It\^o isometry, the accumulated noise has
variance~$\sigman^2(\taun)$.
Writing $\mathbf{x}_{\taun}=\mathbf{x}_0+\sigman(\taun)
\boldsymbol\epsilon$ with
$\boldsymbol\epsilon\sim\mathcal{N}(0,I)$, denoising score matching
\citep{VincentEtAl2011,Song2021SDE} trains a score network
$\mathbf{s}_\theta(\mathbf{x},\taun)\approx\nabla_{\mathbf{x}}
\log p_{\taun}(\mathbf{x})$ by minimizing
\begin{equation}
\mathbb{E}\,\big\|\mathbf{s}_\theta(\mathbf{x}_{\taun},\taun)
  -(\mathbf{x}_0-\mathbf{x}_{\taun})/\sigman^2(\taun)\big\|^2.
\end{equation}
At each fixed $\taun$, the network learns the global score of the
marginal $p_{\taun}$, with adjacent
times coupled through shared parameters. Equation~\eqref{eq:dsm_loss} in \S\,\ref{sec:implementation} gives the $\gn^2$-weighted form used in practice.

\subsection{Optimality within Gaussian surrogates}
\label{sec:optimality}

Among centered, isotropic Gaussian approximations to the physical transport kernel $\Kernel(\taun)$, the moment-matched Gaussian uniquely minimizes cross-entropy. When $\Kernel(\taun)$ has a density and finite differential entropy, this is equivalently the unique minimizer of forward KL. After convolution with $P_0$, the resulting marginal error is controlled by the $W_2$ contraction bound below.

From optimal transport theory \citep{Villani2009}, convolution by an independent variable contracts the $W_2$ distance: for any kernels $\nu_1,\nu_2$,
\begin{equation}
W_2(P_0*\nu_1,\,P_0*\nu_2)\le W_2(\nu_1,\nu_2).
\end{equation}

Let $(\mathbf{Y}_1,\mathbf{Y}_2)$ be an optimal
$W_2$-coupling of $\nu_1$ and~$\nu_2$, and let
$\mathbf{X}_0\!\sim\!P_0$ independently.  Setting
$\mathbf{Z}_i=\mathbf{X}_0+\mathbf{Y}_i$ produces marginals
$P_0*\nu_i$ with
$\|\mathbf{Z}_1-\mathbf{Z}_2\|=\|\mathbf{Y}_1-\mathbf{Y}_2\|$
(the shared~$\mathbf{X}_0$ cancels), so
$W_2^2(P_0*\nu_1,P_0*\nu_2)
  \le\mathbb{E}\|\mathbf{Z}_1-\mathbf{Z}_2\|^2
  =W_2^2(\nu_1,\nu_2)$.
Setting $\nu_1=\Kernel(\taun)$ (physical kernel) and
$\nu_2=\mathcal{N}(0,\sigman^2(\taun)I)$ (Gaussian surrogate
kernel) yields
\begin{equation}
W_2(P^{\mathrm{Phys}}_{\taun},\,p_{\taun})\le
  W_2\!\big(\Kernel(\taun),\,\mathcal{N}(0,\sigman^2(\taun)I)\big),
\end{equation}
so the approximation error is controlled by how far the physical
transport kernel $\Kernel(\taun)$ deviates from Gaussian.

\subsection{Kurtosis evolution and error quantification}
\label{sec:error_quantification}
We quantify higher‑order statistics using excess kurtosis~$K=\kappa_4/\sigma^4$. From the Gaussian convolution structure, the model’s kurtosis obeys the closed‑form evolution given in Eq.~\eqref{eq:kurtosis_evolution_ve_main} of the main text.

For the physical process $P^{\mathrm{Phys}}_{\taun}=P_0*\Kernel(\taun)$, cumulant additivity gives $\kappa_4^{\mathrm{Phys}}=\kappa_4(P_0)+\kappa_4(\Kernel)$, while the VE surrogate has $\kappa_4^M=\kappa_4(P_0)$. We define the kernel's excess kurtosis as $K_{\Kernel}(\taun):=\kappa_4(\Kernel(\taun))/\sigman^4(\taun)$ (subscript $\Kernel$ distinguishes this from the marginal kurtosis $K$). The resulting excess-kurtosis difference is
\begin{equation}
K_{\mathrm{phys}}(\taun)-K_M(\taun)
=K_{\Kernel}(\taun)\left(\frac{\sigman^2(\taun)}{\sigma_0^2+\sigman^2(\taun)}\right)^{\!2}.
\label{eq:kurt_gap}
\end{equation}
The gap vanishes if the kernel Gaussianizes ($K_{\Kernel}\!\to\!0$) and is suppressed early when $\sigman^2\ll\sigma_0^2$.

\subsection{Kac/telegraph kernel: density, moments, and parameter selection}
\label{sec:kac_kernel}

The Kac flow baseline is built on the Goldstein--Kac telegraph process \citep{Goldstein1951,Duong2025Kac}. A particle moves in one dimension at constant speed $c$ and reverses direction at the events of a Poisson process with rate $a$. The initial velocity is $\pm c$ with equal probability. The resulting displacement kernel at time~$t$ is
\begin{align}
p_{\mathrm{K}}(x,t)
&=\frac{e^{-at}}{2}\big[\delta(x-ct)+\delta(x+ct)\big]
\nonumber\\
&\;+\frac{a\,e^{-at}}{2c}
\left[I_0\!\left(\tfrac{a}{c}\eta\right)
+\frac{ct}{\eta}\,I_1\!\left(\tfrac{a}{c}\eta\right)\right]
\mathbf{1}_{\{|x|<ct\}},
\label{eq:kac_density}
\end{align}
where $\eta=\sqrt{c^2t^2-x^2}$ and $I_0$, $I_1$ are modified Bessel functions of the first kind. The kernel has bounded support~$|x|\le ct$.

The kernel moments follow from the velocity autocorrelation $\langle v(t)\,v(t+s)\rangle=c^2 e^{-2as}$. The fourth moment is obtained by integrating $x^4p_{\mathrm K}(x,t)$, including the two point masses in Eq.~\eqref{eq:kac_density}. With the abbreviation $u:=2at$, the kernel variance is
\begin{equation}
\sigma^2_{\mathrm{K}}(t)=\frac{c^2}{2a^2}\left(u-1+e^{-u}\right),
\label{eq:kac_variance}
\end{equation}
and the kernel excess kurtosis, which plays the role of $K_{\Kernel}$ in Eq.~\eqref{eq:kurt_gap}, is
\begin{equation}
K_{\Kernel}^{\mathrm{Kac}}(t)
=3\,\frac{u^2-4u+6-(6+2u)\,e^{-u}}{\left(u-1+e^{-u}\right)^{2}}-3.
\label{eq:kac_kurtosis}
\end{equation}
It equals $-2$ for $t\to0$, where the kernel is a symmetric two-point distribution, and relaxes to zero from below as~$K_{\Kernel}^{\mathrm{Kac}}\simeq-3/(at)$ for $at\gg1$.

The dotted reference curves in Fig.~\ref{fig:variance_kurtosis}b--d show the marginal excess kurtosis of the Kac model. It follows from cumulant additivity for $P_0*p_{\mathrm{K}}$ and combines Eq.~\eqref{eq:kurtosis_evolution_ve_main} with the gap term of Eq.~\eqref{eq:kurt_gap}:
\begin{equation}
K_{\mathrm{Kac}}(\taun)
=\frac{K_0\,\sigma_0^4
+K_{\Kernel}^{\mathrm{Kac}}(\taun)\,\sigman^4(\taun)}
{\left(\sigma_0^2+\sigman^2(\taun)\right)^{2}}.
\label{eq:kac_marginal}
\end{equation}

The parameters $(a,c)$ are not fitted. On the normalized clock, they are fixed by requiring the Kac process to share the anchored telegraph variance path. This gives $a=1/(2\tauc)$ and $c^2=C^{-1}/(2\tauc)$, for which Eq.~\eqref{eq:kac_variance} reduces to Eq.~\eqref{eq:variance_telegraph_norm} with $u=\taun/\tauc$ and $\sigman^2(1)=1$. The Kac baseline is variance-matched to the VE-telegraph schedule by construction, and Table~\ref{tab:ablation_chi2} compares the resulting Gaussian and finite-speed constructions.

The learned Kac-flow baseline follows the conditional-flow-matching construction of \cite{Duong2025Kac}: a bounded velocity network is regressed against the analytical conditional Kac velocity, and sampling integrates the learned flow ODE backward from the terminal marginal.

\subsection{Diffusion model implementation and baseline comparisons}
\label{sec:implementation}

We use the VE framework instead of Variance Preserving (VP/DDPM) models because the transport problems considered here exhibit growing variance. In scattering processes, the spatial variance grows naturally with time ($\sigman^2(\taun)$ increases), matching the VE formulation $d\mathbf{x}_{\taun}=\gn(\taun)\,d\mathbf{W}_{\taun}$, with
$\gn^2(\taun)=d\sigman^2(\taun)/d\taun$. In contrast, VP models constrain variance to unity ($\sigma^2=1$) via a mean-reverting drift, which would obscure the direct physical interpretation of the transport coordinates. For finite-speed transport, $\sigman^2(\taun)$ follows the normalized telegraph law (Eq.~\ref{eq:variance_telegraph_norm} in the main text). Baseline schedules use linear $\sigman^2(\taun)\propto\taun$ (pure diffusion at all $\taun$), cosine schedules \citep{NicholDhariwal2021}, and a Karras/EDM-inspired power-law variance schedule $\sigman^2(\taun)\propto\taun^\rho$ with $\rho=7$ \citep{Karras2022EDM}. All models share the same score network architecture (a 10-block FiLM-MLP with GroupNorm, using sinusoidal time embeddings~\citep{Song2021SDE} and FiLM conditioning; approximately $1.16\times10^{6}$ learnable parameters). The physics-anchored variant fixes~$\sigman^2(\taun)$ via the physical law; architecture and computational cost are unchanged---only the scalar schedule~$\gn^2(\taun)$ differs.

The score network optionally conditions on entrance kurtosis $K_0$ through a small MLP that maps $\log(K_0)$ (normalized to zero mean and unit variance) to a 32-dimensional embedding, which modulates hidden features via FiLM layers alongside the 64-dimensional sinusoidal time embedding. This enables a single trained model to serve multiple entrance distributions without retraining.

The complete training and sampling procedure is summarized in Algorithm~\ref{alg:pa_ve}; all hyperparameters and architecture details are listed in Table~\ref{tab:compute}. Training minimizes the~$\gn^2$-weighted denoising score-matching loss,
\begin{equation}
  \mathcal{L}(\theta)=\mathbb{E}_{\taun,\mathbf{x}_0,\boldsymbol\epsilon}\!\Big[\,
    \gn^{2}(\taun)\;\big\lVert\,
      \mathbf{s}_\theta(\mathbf{x}_\taun,\taun)
      -(\mathbf{x}_0-\mathbf{x}_\taun)/\sigman^2(\taun)
    \,\big\rVert^{2}\Big],
  \label{eq:dsm_loss}
\end{equation}
with $\mathbf{x}_\taun=\mathbf{x}_0+\sigman(\taun)\boldsymbol\epsilon$ and $\boldsymbol\epsilon\sim\mathcal{N}(\mathbf{0},I)$. The $\gn^2$ prefactor is standard signal-to-noise weighting \citep{Song2021SDE,Karras2022EDM}; it upweights time steps where the schedule is steepest, ensuring the network allocates capacity where the marginal changes fastest.

\begin{table}[t]
\centering
\caption{\textbf{Training and sampling configuration.} All VE variants share identical architecture and budget; only the schedule $\gn^2(\taun)$ differs. The Kac flow baseline uses the same architecture with a bounded output layer ($c\!\cdot\!\tanh$).}\label{tab:compute}
\begin{tabular}{@{}ll@{}}
\toprule
\textbf{Item} & \textbf{Value} \\
\midrule
\multicolumn{2}{@{}l}{Architecture} \\
\quad Score network          & 10-block FiLM-MLP, GroupNorm \\
\quad Hidden channels $h$    & 256 \\
\quad Time embedding dim     & 64 (sinusoidal) \\
\quad Kurtosis embedding dim & 32 (MLP + FiLM) \\
\quad Learnable parameters   & $\approx 1.16\times 10^{6}$ \\
\midrule
\multicolumn{2}{@{}l}{Training} \\
\quad Entrance data           & Student-$t$ samples, $\sigma_0^2=0.08$ \\
\quad Entrance kurtoses $K_0$ & $\{0.5,\, 1.0,\, 1.5\}$ (excess) \\
\quad Batch size              & 32 \\
\quad Optimizer / LR          & AdamW / $10^{-3}$ \\
\quad Gradient clipping       & max-norm $1.0$ \\
\quad EMA decay               & $0.999$ \\
\quad Epochs $\times$ iters   & $200 \times 200 = 40\,000$ steps \\
\midrule
\multicolumn{2}{@{}l}{Sampling} \\
\quad PF-ODE solver      & Euler (first-order) \\
\quad Discrete steps $T$   & 500 \\
\quad Requested readout points & 40, log-spaced in $[3 \times 10^{-3},\,1]$ \\
\quad Samples per readout & $256\times32\times32=262{,}144$ scalars \\
\quad Seeds (uncertainty) & 10 \\
\midrule
\multicolumn{2}{@{}l}{Compute} \\
\quad Hardware              & single NVIDIA A100 (40\,GB) \\
\quad Training wall-clock   & ${\approx}\,10$\,min per model \\
\quad Sampling (256, 40 readouts) & ${\approx}\,1$\,min \\
\bottomrule
\end{tabular}
\end{table}

At inference we draw terminal states from the exact forward marginal $p_1$ implied by the VE process (i.e.\ $\mathbf{x}_1=\mathbf{x}_0+\tilde\sigma(1)\boldsymbol\epsilon$ with $\mathbf{x}_0\sim P_0$) and integrate the PF-ODE backward with an Euler solver ($T=500$ steps), halting at the desired readout $\taustar$. Uncertainty is quantified over 10 independent seeds.

% ── Algorithm 1 ──────────────────────────────────────────────
\begin{algorithm}[t]
\caption{Physics-anchored VE diffusion: training and sampling.}\label{alg:pa_ve}
\begin{algorithmic}[1]

% ── INPUTS ──
\Statex \textbf{Input:}
  Variance path $\sigman^2(\taun)$ from physical law, with $\sigman^2(1)=1$;
  entrance samples $\mathbf{x}_0\!\sim\!P_0$
\Statex \textbf{Define:}
  $\gn^2(\taun)=d\sigman^2/d\taun$; \quad
  $\sigman(\taun)=\sqrt{\smash[b]{\sigman^2(\taun)}}$

\Statex
\Statex \hrulefill
\Statex \textbf{Training} \hfill (entrance-only denoising score matching)
\Statex \hrulefill

\Repeat
  \State Sample $\mathbf{x}_0\!\sim\!P_0$
  \State Sample $\taun\!\sim\!\mathcal{U}(0,1]$
  \State Sample $\boldsymbol\epsilon\!\sim\!\mathcal{N}(\mathbf{0},I)$
  \State $\mathbf{x}_\taun\leftarrow\mathbf{x}_0+\sigman(\taun)\,\boldsymbol\epsilon$
  \Comment{Noisy state at physical time $\taun$}
  \State $\mathcal{L}\;\leftarrow\;
    \gn^{2}(\taun)\;\big\lVert\,
      \mathbf{s}_\theta(\mathbf{x}_\taun,\taun)
      -(\mathbf{x}_0-\mathbf{x}_\taun)/\sigman^2(\taun)
    \,\big\rVert^{2}$
  \Comment{$\gn^2$-weighted loss}
  \State Update $\theta$ via AdamW on $\nabla_\theta\mathcal{L}$;
         \; update EMA shadow
\Until{converged}

\Statex
\Statex \hrulefill
\Statex \textbf{Sampling} \hfill (physics-timed PF-ODE at readout $\taustar$)
\Statex \hrulefill

\State Draw $\mathbf{x}_0\!\sim\!P_0$, $\boldsymbol\epsilon\!\sim\!\mathcal{N}(\mathbf{0},I)$; set $\mathbf{x}_1\leftarrow\mathbf{x}_0+\sigman(1)\,\boldsymbol\epsilon$
\Comment{Terminal state from $p_1$}
\State Integrate backward from $\taun\!=\!1$ to $\taun\!=\!\taustar$:
\Statex \qquad $\displaystyle
  \frac{d\mathbf{x}}{d\taun}
  =-\tfrac{1}{2}\,\gn^{2}(\taun)\,
    \mathbf{s}_\theta(\mathbf{x},\taun)$
\State \Return $\mathbf{x}_{\taustar}\sim p_{\taustar}$
\Comment{Anchored surrogate marginal at detector time}
\end{algorithmic}
\medskip
\noindent\footnotesize
\textbf{Key point:} only the scalar schedule $\gn^2(\taun)$ changes relative to a standard VE model; the network architecture, optimizer, and sampler are unchanged. Optionally, the score network conditions on entrance kurtosis $K_0$ via a FiLM embedding (see text), enabling a single model for multiple entrance distributions.
\end{algorithm}

\subsection{Likelihood computation and differentiable inference}
\label{sec:likelihood_fisher}
The probability-flow ODE corresponding to Eq.~\eqref{eq:ve_forward_main} is
\begin{equation}
\frac{d\mathbf{x}}{d\taun}= -\tfrac{1}{2}\gn^2(\taun)\,\mathbf{s}_\theta(\mathbf{x},\taun).
\end{equation}
The instantaneous change-of-variables formula \citep{Song2021SDE} yields the PF‑ODE model log‑likelihood:
\begin{equation}
\log p_{\taustar}(\mathbf{x}_{\taustar})=\log p_1(\mathbf{x}_1)-\int_{\taustar}^{1}\!\tfrac{1}{2}\gn^2(\taun)\,\mathrm{div}\,\mathbf{s}_\theta(\mathbf{x}_{\taun},\taun)\,d\taun,
\label{eq:exact_likelihood_methods}
\end{equation}
where the divergence is estimated using Hutchinson's trace estimator \citep{Hutchinson1990}. This expression yields the exact marginal log-likelihood in the limit where $\mathbf{s}_\theta$ matches the true score and the terminal density $p_1$ is specified; otherwise it provides a differentiable likelihood estimate. Physical parameters that enter the schedule (e.g., $\tauc$) may be included in $\sigman^2(\taun;\tauc)$, while entrance parameters such as $K_0$ may parameterize~$P_0(\,\cdot\,;K_0)$ or condition $\mathbf{s}_\theta$; such parameters can then be estimated by maximizing Eq.~\eqref{eq:exact_likelihood_methods}.

\subsection{Charged-particle transport through magnetized turbulence}\label{sec:CR_transport}

Throughout this paper, $\taun\in[0,1]$ denotes normalized time. Physical time $t$ relates to $\taun$ via the total observation window (the normalized correlation time is $\tauc=t_c/t_{\mathrm{obs}}$ where $t_c=\ell_c/v$ is the cell-crossing time).

The transverse acceleration of a charged particle along its path through magnetized turbulence $\mathbf{B}$ is
\begin{equation}
a(t)\;\equiv\;\frac{d v_\perp}{dt}\;\simeq\;\frac{q\,v}{m}\,B_\perp.
\end{equation}
Since $v_\perp(t)=\int_0^t a_\perp(s)\,ds$, the velocity
variance is a double integral of the acceleration autocorrelation:
\begin{equation}
\sigma^2_{v_\perp}(t)
  \;=\;\int_0^t\!\!\int_0^t\! C_a(s{-}s')\,ds\,ds'
  \;=\;2\!\int_0^t\!(t{-}\Delta t)\,C_a(\Delta t)\,d\Delta t.
\label{eq:taylor_green_kubo}
\end{equation}
For $t\!\ll\!t_c$ the autocorrelation is nearly constant, $C_a\!\approx\!\langle a^2\rangle$, giving $\sigma^2_{v_\perp}\!\approx\!\langle a^2\rangle\,t^2$ (ballistic). For $t\!\gg\!t_c$ the factor $(t{-}\Delta t)\!\approx\!t$ over the correlation width and the integral saturates, giving $\sigma^2_{v_\perp}\!\approx\!2\,\Dvel\,t$ (diffusive), where
\begin{equation}
\Dvel\;=\;\int_0^\infty C_a(\Delta t)\,d\Delta t\;\approx\;\langle a^2\rangle\,\frac{\ell_c}{v}.
\end{equation}
Equivalently, from a random‑walk estimate one obtains~$\Dvel \simeq q^2 B_{\mathrm{rms}}^2\,\ell_c\,v/(2m^2)$. Here, we used the Taylor-Green–Kubo formula (see \citep{Kubo_1957}, and references therein) and $C_a(\Delta t)\;:=\;\big\langle a_\perp(t)\,a_\perp(t+\Delta t)\big\rangle$ is the stationary autocorrelation of the transverse acceleration $a_\perp(t)$.

In physical time $t$ (with correlation time $t_c = \lc/v$), the expected crossovers follow:
\begin{equation}
\sigma^2_{v_\perp}(t)\ \propto\
\begin{cases}
\langle a^2\rangle\,t^2, & t\ll t_c \quad\text{(ballistic)},\\[2pt]
\Dvel\,t, & t\gg t_c \quad\text{(diffusive)}.
\end{cases}
\end{equation}
The corresponding displacement scaling in the telegraph (finite-speed) transport picture is:
\begin{equation}
\sigma^2_{x_\perp}(t)\ \propto\
\begin{cases}
t^2, & t\ll t_c \quad\text{(ballistic)},\\[2pt]
\kappaS\,t, & t\gg t_c \quad\text{(diffusive)},
\end{cases}
\end{equation}
Upon normalizing $\taun = t/t_{\mathrm{obs}}$, these give the short- and long-time limits of Eq.~\eqref{eq:variance_telegraph_norm}.
Here, \(\kappaS\) is the spatial diffusion coefficient, $\langle a^2\rangle$ is absorbed into~$\Dvel$ via the Green--Kubo integral ($\Dvel\approx\langle a^2\rangle\,\ell_c/v$), and $\kappaS=v^4/(6\Dvel)$ absorbs all remaining acceleration dependence into a single transport coefficient.

Note that \(\kappaS\) can be inferred in two equivalent ways: (i) from the long-time slope $\sigma^2_{x_\perp}(t)=2\kappaS\,t$; or (ii) from the measured \(D_v\) via small‑angle scattering,
\begin{equation}
\kappaS=\frac{v^2}{3\,\nu}=\frac{v^4}{6\,\Dvel},
\end{equation}
where $\nu$ is the pitch‑angle scattering rate.

\subsection{Kurtosis inheritance and relaxation in turbulent transport}\label{sec:methods_kurtosis_relaxation}
In the high-rigidity limit ($r_g\gg\ell_c$), transport proceeds via independent scattering events as particles traverse turbulent cells. Each signed transverse-component kick is proportional to the path-averaged transverse field $\overline{B}_\perp$:
\begin{equation}
\Delta v_\perp(\tauc)\propto\overline{B}_\perp.
\end{equation}
This linear relationship preserves kurtosis: $K(\Delta v_\perp(\tauc))=K(\overline{B}_\perp)$.

After accumulating $N=\ell/\ell_c$ approximately independent kicks, cumulant additivity gives $\kappa_4(v_\perp^{\mathrm{total}})=N\,\kappa_4(\overline{B}_\perp)$, while variance scales as $\sigma^2=N\,\sigma^2_{\mathrm{kick}}$. Since excess kurtosis is normalized by variance squared ($K=\kappa_4/\sigma^4$),
\begin{equation}
K\big(v_\perp^{\mathrm{total}}\big)
=\frac{N\,\kappa_4(\overline{B}_\perp)}{(N\,\sigma^2_{\mathrm{kick}})^2}
=\frac{1}{N}\,K(\overline{B}_\perp),
\label{eq:kurtosis_inheritance_methods}
\end{equation}
describing the slow Gaussianization observed as the distribution propagates through the turbulent medium (Fig.~\ref{fig:propagation_cartoon}f). This $1/N$ decay holds once transport has entered the decorrelated regime ($\ell\gtrsim\ell_c$). For $\ell\lesssim\ell_c$ (the ballistic/crossover regime), finite-speed effects can make the true kernel non-Gaussian; in that regime the VE construction serves as a variance-correct Gaussian surrogate, with kernel mismatch quantified by Eq.~\eqref{eq:kurt_gap}.

It is instructive to compare this physical $1/N$ law with the surrogate. Setting $\sigma_0^2=\sigma^2_{\mathrm{kick}}$ and $\sigman^2=(N{-}1)\,\sigma^2_{\mathrm{kick}}$ in Eq.~\eqref{eq:kurtosis_evolution_ve_main} gives $K_M=K_0/N^2$: the surrogate carries only the entrance contribution to $\kappa_4$ and therefore relaxes one power of $N$ faster than Eq.~\eqref{eq:kurtosis_inheritance_methods}. The kurtosis gap restores the difference exactly: the kick-accumulation kernel has $K_{\Kernel}=K_0/(N{-}1)$, for which Eq.~\eqref{eq:kurt_gap} evaluates to $K_0(N{-}1)/N^2$, so that $K_M$ plus the gap reproduces the physical $K_0/N$. This component-wise comparison is evaluated directly in Fig.~\ref{fig:propagation_cartoon}f.

\subsection{MHD-grid turbulence (periodic cube)}
\label{sec:methods_mhdgrid}

We use a pre-computed realization of a three-dimensional solenoidal magnetic field from the Johns Hopkins Turbulence Databases \citep{Eyink_2013Natur.497..466E, JHTDB} on a uniform periodic grid of size $N^3$ ($N=1024$). The vector field is then loaded into \texttt{CRPropa} within a \texttt{MagneticFieldGrid} \citep{CRPropa_2022JCAP...09..035A}. The box size is set from the integral scale quoted for this dataset, $\ell_{\rm int}=0.217\,L_{\rm box}$, so that $L_{\rm box}=\lc/0.217=461\,\mu$m for $\lc=100\,\mu$m. The field is rescaled so that the mean field magnitude $\langle|\mathbf{B}|\rangle$ equals the target value (for this realization $B_{\rm rms}/\langle|\mathbf{B}|\rangle=1.17$). Reference parameters are listed in Table~\ref{tab:baseline_params}.

\begin{table*}[t]
\centering
\caption{\textbf{Reference parameters for the turbulence benchmark and detector geometry.} Experimental values are taken from \cite{Chen2020} where indicated.}
\label{tab:baseline_params}
\resizebox{\linewidth}{!}{%
\begin{tabular}{l l l l}
\toprule
\textbf{Quantity} & \textbf{Symbol} & \textbf{Reference value} & \textbf{Notes / Source} \\
\midrule
RMS magnetic field & $B_{\rm rms}$ & $\sim100$\,kG  & Proton radiography \& FLASH. \\
& & &\citep{Chen2020} \\
Outer/driving scale & $L$ & $\sim 400~\mu$m $\approx 0.35\,L_{\rm box}$ & Set by offset grids. \citep[Fig.\,1c]{Chen2020} \\
Correlation length & $\ell_c$ & $50$--$90~\mu$m & Lab inference. \citep[][App.~B]{Chen2020} \\
Spectrum & $E_B(k)$ & inherited from dataset & \\
Dissipation scale & $\ell_{\min}$ & a few grid cells & Chosen to match dissipation range. \\
& & & \citep[Fig.\,1c]{Chen2020} \\
Domain extent & $L_{\rm dom}$ & $L_{\rm box}$ (periodic) &  \\
Pinhole radius & $R_{\rm pin}$ & $\sim100~\mu$m & Lab geometry. \citep[Fig.\,3]{Chen2020} \\
Detector distance & $r_{\rm det}$ & $27$\,cm & Lab geometry. \citep[][§2]{Chen2020} \\
Ring statistic &  & largest closed 20\% contour & Defined w.r.t.\ mean flux inside $R_{\rm pin}$. \\
& & & \citep[][App.~C.2]{Chen2020} \\
Ring $\to$ velocity &  & $\Delta v_\perp=(1.65/r_{\rm det})\,\Delta r_{\rm rms}\,v$ &  Smearing model + geometry. \\
& & & \citep[][Eqs.\ C10,C20]{Chen2020} \\
\bottomrule
\end{tabular}%
}
\end{table*}

\subsection{Swiss roll ablation and $k$‑NN P/R analysis}\label{sec:methods_swiss_pr}
We evaluate schedule choice on a 2-D swiss roll manifold. The score network is trained once with $\sigma$-conditioning (the network learns score functions at many noise levels~$\sigma$ simultaneously, without committing to a specific time-to-noise mapping) using $\sigma\sim\mathrm{LogNormal}(\mu=-1,\sigma=1)$ and sinusoidal embeddings. At sampling, we vary only the path~$\sigma(\tau)$---Telegraph, Karras EDM, Linear, or Cosine---using identical Karras-Heun ODE solvers, NFE, seeds, and noise bounds \citep{Karras2022EDM,NicholDhariwal2021}.

For evaluation, we draw $M$ ground-truth points from the swiss roll and $M$ generated points (same $M$ for all schedules). We repeat sampling with five independent seeds and report mean $\pm 1\sigma$. For each neighborhood size $k\in\{1,\dots,50\}$, we compute the nearest-center precision/recall diagnostic defined below, adapted from~\cite{Kynkaanniemi2019}.

Let $r_{\mathrm{data}}(x_i)$ be the distance from data point $x_i$ to its $k$-th nearest neighbor in the real set, and let $d_{\mathrm{g\to d}}(\hat{x}_j)$ be the distance from generated point $\hat{x}_j$ to its nearest real neighbor. Precision measures what fraction of generated samples lie within the k-NN bubble of their nearest real neighbor:
\begin{equation}
\mathrm{Prec}_k=\frac{1}{M}\sum_{j=1}^M\mathbf{1}\!\left[d_{\mathrm{g\to d}}(\hat{x}_j)\le r_{\mathrm{data}}\!\big(\mathrm{NN}_{\mathrm{data}}(\hat{x}_j)\big)\right].
\end{equation}
High precision means generated samples stay close to the real manifold (no spurious modes). Similarly, let $r_{\mathrm{gen}}(\hat{x}_j)$ be the distance from generated point $\hat{x}_j$ to its $k$-th nearest generated neighbor, and $d_{\mathrm{d\to g}}(x_i)$ the distance from real point~$x_i$ to its nearest generated neighbor. Recall measures what fraction of real samples have a generated neighbor within their k-NN bubble:
\begin{equation}
\mathrm{Rec}_k=\frac{1}{M}\sum_{i=1}^M\mathbf{1}\!\left[d_{\mathrm{d\to g}}(x_i)\le r_{\mathrm{gen}}\!\big(\mathrm{NN}_{\mathrm{gen}}(x_i)\big)\right].
\end{equation}
High recall means the generator covers all regions of the real distribution (no missing modes). Small $k$ tests strict local fidelity (sharpness), while large $k$ tests global structure (coverage). 
At very local scales (small $k$), EDM attains high recall quickly (broad coverage) but at the cost of precision (lower fidelity), consistent with early aggressive denoising. Telegraph strikes a better local balance. Across five independent sampling seeds, the telegraph path attains higher recall than cosine at small $k$ at comparable precision and maintains higher precision than EDM at every neighborhood size, while cosine retains a modest precision lead at intermediate $k$; telegraph and cosine both approach high precision and near-unit recall for $k\!\gtrsim\!40$. Quantitatively, at $k{=}5$, telegraph attains recall $0.83$ vs.\ $0.72$ for cosine at comparable precision ($0.43$ vs.\ $0.44$); at $k{=}20$, cosine leads in precision ($0.91$ vs.\ $0.86$). The linear schedule has precision $<0.16$ for $k\!\le\!20$.

\section*{Data availability}
The source data underlying all figures and tables, together with the pre-computed simulation and ablation outputs generated in this study, have been deposited in Zenodo~\citep{ReichherzerZenodo2026}. The forced-MHD turbulence field used in this study is available from the Johns Hopkins Turbulence Database under DOI \url{https://doi.org/10.7281/T1930RBS}. The experimental comparison data were taken from Ref.~\cite{Chen2020}.

\section*{Code availability}
The code used to generate the results and figures has been deposited in Zenodo~\citep{ReichherzerZenodo2026} and is available under the MIT License. Test-particle simulations use the open-source \texttt{CRPropa} framework~\citep{CRPropa_2022JCAP...09..035A}.

\section*{Acknowledgments}
We thank Archie F. A. Bott, Robert J. Ewart, and Jonathan Citrin for valuable comments. 

\section*{Funding}
The work of P.R. was funded  through the Walter Benjamin Fellowship by the Deutsche Forschungsgemeinschaft (DFG, German Research Foundation)---grant numbers 518672034 and 593157299; P.R. was also supported by a postdoc
fellowship of the German Academic Exchange Service (DAAD); The work of G.G. was partially supported by UKRI under grants no. ST/W000903/1 and EP/Y035038/1.

\section*{Author contributions statement}
P.R.\ conceived the method, developed the theoretical framework, wrote all code, performed the simulations and analysis, and wrote the original manuscript. D.N.H.\ refined the theoretical formulation and substantially revised the manuscript's narrative, structure, and presentation. G.G.\ and S.S.\ provided scientific guidance and contributed to the interpretation of the laboratory validation through their expertise in the experimental data and associated uncertainties. All authors discussed the results and reviewed and edited the manuscript.

\section*{Competing interests statement}
The authors declare no competing interests.

\bibliography{references}
\bibliographystyle{sn-basic}

\end{document}